\documentclass{article}
\usepackage{ijcai20}

\usepackage{times}
\usepackage{soul}
\usepackage{url}
\usepackage[hidelinks]{hyperref}
\usepackage[utf8]{inputenc}
\usepackage[small]{caption}
\usepackage{graphicx}
\usepackage{amsmath}
\usepackage{amsthm}
\usepackage{amsfonts}
\usepackage{amssymb}
\usepackage{graphicx}
\usepackage{bm}
\usepackage{booktabs}
\usepackage{algorithm}
\usepackage{algorithmic}
\usepackage{multirow}
\usepackage{multicol}
\usepackage{footnote}
\usepackage{color}

\title{On the Importance of Word and Sentence Representation Learning in \\ Implicit Discourse Relation Classification}

\author{
Xin Liu$^1$\footnote{This work was done when Xin Liu was an intern at Huawei Noah's Ark Lab.}
\and
Jiefu Ou$^1$\and
Yangqiu Song$^1$\And
Xin Jiang$^2$
\affiliations
$^1$Department of CSE, the Hong Kong University of Science and Technology\\
$^2$Huawei Noah’s Ark Lab\\
\emails
xliucr@cse.ust.hk,
jouaa@connect.ust.hk,
yqsong@cse.ust.hk,
jiang.xin@huawei.com
}

\begin{document}

\maketitle

\begin{abstract}
  Implicit discourse relation classification is one of the most difficult parts in shallow discourse parsing as the relation prediction without explicit connectives requires the language understanding at both the text span level and the sentence level.
  Previous studies mainly focus on the interactions between two arguments. 
  We argue that a powerful contextualized representation module, a bilateral multi-perspective matching module, and a global information fusion module are all important to implicit discourse analysis.
  We propose a novel model to combine these modules together.
  Extensive experiments show that our proposed model outperforms BERT and other state-of-the-art systems on the PDTB dataset by around 8\% and CoNLL 2016 datasets around 16\%. 
  We also analyze the effectiveness of different modules in the implicit discourse relation classification task and demonstrate how different levels of representation learning can affect the results.
\end{abstract}
\section{Introduction}

Discourse parsing is a fundamental task in natural language processing, which is considered to be a crucial step for many downstream tasks, such as machine translation~\cite{Li2014Assessing}, question answering~\cite{Jansen2014Discourse}, text generation~\cite{Bosselut2018Discourse}, and text summarization~\cite{Cohan2018A}.
Shallow discourse parsing, such as Penn Discourse Treebank 2.0 (PDTB 2.0)~\cite{Prasad2008The} and  CoNLL shared task~\cite{Xue2016CoNLL}, focuses on identifying relations between two text spans (i.e., clauses or sentences) that are named discourse units.
Among different sub-tasks in PDTB and CoNLL shared task, implicit relation classification between two sentences, with the absence of discourse markers or connectives in between (e.g., however, because), is considered as the most difficult problem.

Machine learning has been widely used in implicit discourse relation classification.
Traditional research mainly focuses on linguistic features generation and selection at the text span level and sentence level, and the chosen classifiers are often linear classifiers~\cite{Pitler2009Automatic,Lin2009Recognizing}. 
Recently, more flexible deep neural architectures are proposed to learn better representations. 
A considerable amount of effort has also been put for finding a good neural network architecture for implicit relation classification~\cite{Zhang2015Shallow,Liu2016Implicit,Qin2017Adversarial,Bai2018Deep,Dai2018Improving} or some better learning settings such as multi-task learning~\cite{Liu2016Implicit,Lan2017Multi,Varia2019Discourse}.
More recently, contextualized word representation learning, such as ELMo~\cite{Peters2018Deep}, BERT~\cite{Devlin2019BERT}, and RoBERTa~\cite{Liu2019RoBERTa}, has been proven to significantly improve many downstream tasks.
It has been shown that BERT, especially the {\it next sentence prediction} (NSP) task, can significantly improve the implicit discourse relation classification task~\cite{Shi2019Next}.
However, in RoBERTa, while using a larger amount of data training and achieving better performance on many different tasks, it is claimed that NSP is not used.
Nonetheless, we find RoBERTa cannot surpass BERT in this task without great modifications. 
Thus, it is unclear how to choose a contextualized encoder and what leads to improvement.


In this work, we show that different levels of representation learning are all important to this difficult implicit relation classification problem: deep contextualized word representations associated with an additional sentence separator can improve argument representations by better integrating context information; a bilateral multi-perspective matching can learn the interaction between two arguments without the heavy burden of feature engineering; a fusion module combining the multi-head attention and the gate mechanism can directly collect important clues to understand the text in depth.
To this end, we propose a novel model, BMGF-RoBERTa (\textbf{B}ilateral \textbf{M}atching and \textbf{G}ated \textbf{F}usion with \textbf{RoBERTa}), to combine these modules together to address the discourse relation recognition with the absence of explicit connectives.
Extensive experiments show that our proposed model outperforms BERT and other state-of-the-art systems on the standard PDTB dataset around 8\% and CoNLL datasets around 16\%.
We also systematically analyze the effectiveness of different modules and demonstrate how different levels of semantic representation learning contribute to the discourse relation recognition problem. The source code is available at: \url{https://github.com/HKUST-KnowComp/BMGF-RoBERTa}.

\section{Related Work}

\begin{figure}[t]
\centering
\includegraphics[width=0.9\linewidth]{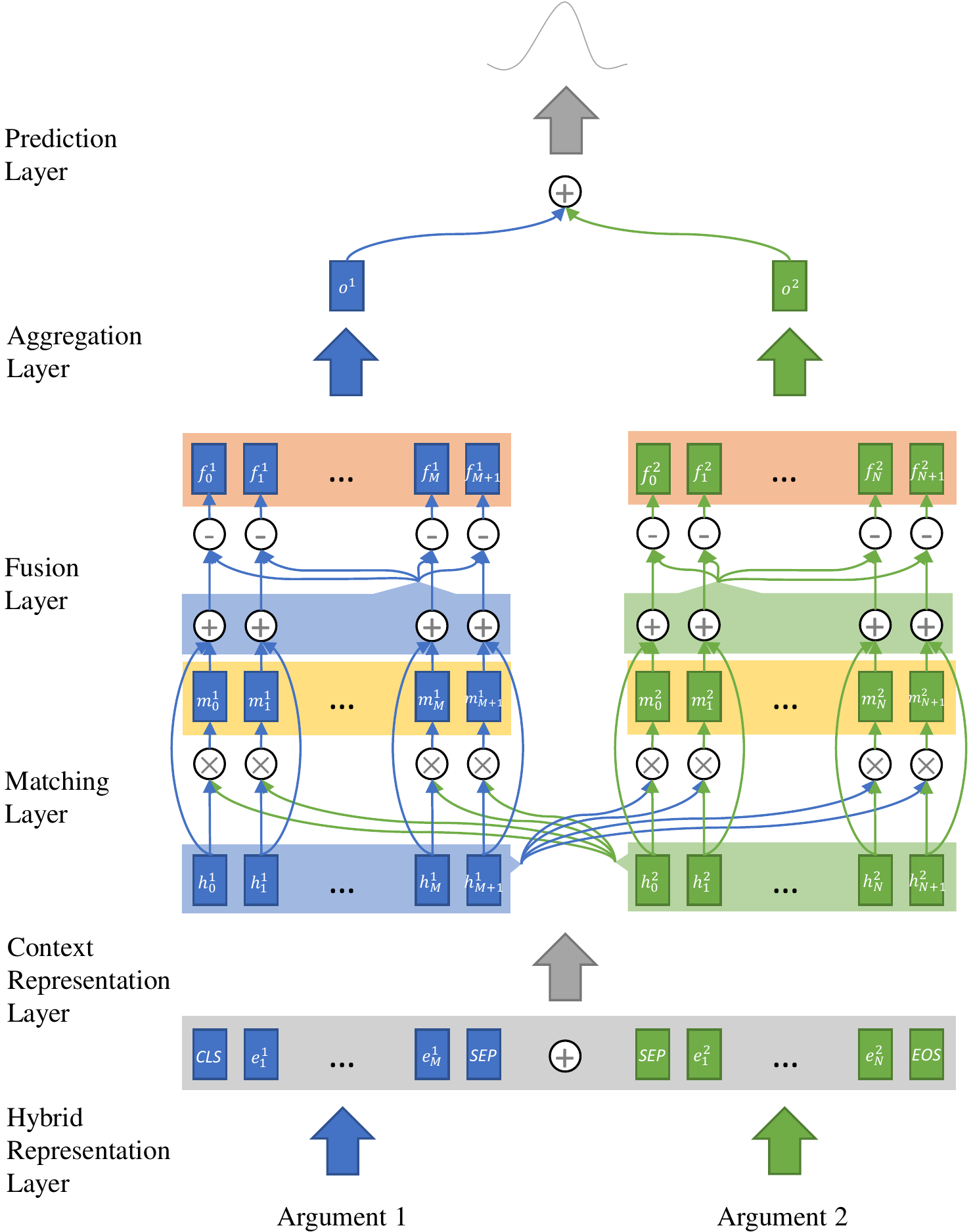}
\caption{The Architecture Overview.}
\label{fig:arch}
\end{figure}

After the release of Penn Discourse Treebank 2.0 (PDTB 2.0)~\cite{Prasad2008The}, the task of shallow discourse parsing has received much attention.
Despite the great progress in classifying explicit discourse relations with F1 scores higher than 93\% on four-class classification~\cite{Pitler2008Easily} and 90\% on 15-class classification~\cite{Wang2015A}, implicit discourse relation recognition still remains a challenge. 
Most approaches regard the recognition problem as a classification problem of pairs of arguments.
Previous research mainly focuses on feature engineering for linear classifiers to classify implicit discourse relations.
The first work~\cite{Pitler2009Automatic} selected several surface features to train four binary classifiers for the four top-level classes on PDTB.
Results showed word pairs were most important among all features developed.
\cite{Lin2009Recognizing} was the first work to consider the classification on second level types.
They further employed four different feature types to represent the context, the constituent parse trees, the dependency parse trees, and the raw text.
\cite{Rutherford2014Discovering} improved the baseline performance by as much as 25\% with the help of brown cluster pairs and coreference patterns.

Recently, with the development of deep learning, more and more researchers choose to design neural architectures for this task as well as other NLP tasks.
\cite{Ji2015One} adopted two recursive neural networks to exploit the representation of arguments and entity spans.
\cite{Liu2016Recognizing} introduced the multi-level attention to mimic the repeated reading strategy and extract efficient features dynamically.
\cite{Qin2016A} found stacking gated convolutional neural networks can further improve the performance, compared with MLP, LSTM, and highway networks.
\cite{Bai2018Deep} proposed a model with different grained text representations, including character-, subword-, word-, sentence-, and sentence pair- levels, to enhance the representation power.
\cite{Dai2018Improving} took the paragraph as a whole to and modeled inter-dependencies of discourse units by a conditional random filed layer.
\cite{Varia2019Discourse} introduced word/n-gram pair convolutions with a gating mechanism to capture the interaction between the arguments for either explicit or implicit relation classification.
\cite{Shi2019Next} found that the next sentence prediction played a role in finding relationships between sentences and benefited to the implicit discourse relation classification task.
Giant contextualized representation models with pretraining show their power in many natural language understanding tasks even those low-resource tasks~\cite{Peters2018Deep,Devlin2019BERT,Liu2019RoBERTa}.
We can also easily find the trend to work with them from the latest research work in implicit relation classification, e.g., ELMo~\cite{Dai2019A} and BERT~\cite{Shi2019Next}. 

Some researchers chose different routines by expanding training data or integrating human-selected knowledge.
\cite{Liu2016Implicit,Lan2017Multi} shared word embeddings with explicit relation classification.
\cite{Xu2018Using} used active learning to sample explicitly-related arguments and expanded informative and reliable instances.
External event knowledge and coreference relations in the paragraph-level were incorporated into discourse parsing by \cite{Dai2019A}.

\section{Model}
In this section, we introduce the details of our model.
\subsection{Model Overview}

Figure~\ref{fig:arch} shows the architecture of our model.
Our model consists of six parts, including (1) a hybrid representation layer that maps each word to a hybrid between character- and word-level embedding, (2) a contextualized representation layer that enhances the representation power of embeddings, (3) a matching layer that compares each token of one argument against all tokens of the other one and vice versa, (4) a fusion layer that assigns different importance to each word in arguments based on arguments themselves as well as matching results, (5) an aggregation layer that aggregates fusion results and encodes each argument into a vector, (6) a prediction layer that evaluates the probability distribution $\text{P}(y|Arg_1, Arg_2)$.
Finally, we name our model as BMGF-RoBERTa (\textbf{B}ilateral \textbf{M}atching and \textbf{G}ated \textbf{F}usion with \textbf{RoBERTa}).

\subsection{Contextualized Argument-aware Representation}

Contextualized word representations have helped many models improve their performance in different natural language processing tasks.
Some encoders, e.g., ELMO~\cite{Peters2018Deep}, BERT~\cite{Devlin2019BERT}, and RoBERTa~\cite{Liu2019RoBERTa}, have been widely applied.
Here, we use RoBERTa since it is reported to have the best performance on various tasks.

We first use a byte-level Byte-Pair Encoding (BPE) in RoBERTa to encode arguments and map encoded results into hybrid representations:
\begin{align}
    Arg_1&: [\bm{CLS}, \bm{e}_1^1, \bm{e}_2^1, \cdots, \bm{e}_M^1, \bm{EOS}] \nonumber \\
    Arg_2&: [\bm{CLS}, \bm{e}_1^2, \bm{e}_2^2, \cdots, \bm{e}_N^2, \bm{EOS}],
\end{align}
where $\bm{CLS}$ and $\bm{EOS}$ are two special token embeddings introduced in BERT and RoBERTa for attention and downstream tasks. $M$ is the length of $Arg_1$ and $N$ is the length of $Arg_2$, and $\bm{e}_i^j$ is the hybrid between character- and word-level embedding of the $j^{th}$ word in $Arg_i$.
Then the embedding lengths of $Arg_1$ and $Arg_2$ respectively are $M+2$ and $N+2$.

Then we concatenate the token representations as follows:
\scalebox{0.925}{\parbox{1.08\linewidth}{
\begin{align}
    &[\bm{e}_0, \bm{e}_1, \cdots, \bm{e}_M, \bm{e}_{M+1}, \bm{e}_{M+2}, \bm{e}_{M+3}, \cdots, \bm{e}_{M+N+2}, \bm{e}_{M+N+3}] \nonumber \\
    &=[\bm{CLS}, \bm{e}_1^1, \cdots, \bm{e}_M^1, \bm{SEP}, \bm{SEP}, \bm{e}_1^2, \cdots, \bm{e}_N^2, \bm{EOS}],
\end{align}
}}
where $\bm{SEP}$ is a special token embedding to indicate boundaries of the sentence concatenation. We use two $\bm{SEP}$s here since we will further split these embeddings into two arguments and perform matching and fusion shown in the following sections.

Position embeddings are necessary for transformer-based models as shown in BERT and RoBERTa because the order and the distance between two words affect their dependence.
Moreover, RoBERTa does not use segment embeddings to show the order of two sentences after removing the {\it next sentence prediction} (NSP) loss.
However, as shown in \cite{Shi2019Next}, the NSP task is very helpful.
Therefore, we add back the trainable segment embeddings as what NSP does in addition to the hybrid embeddings and the position embeddings.

After multiple transformer layers, we get the contextualized representations $[\bm{h}_0, \bm{h}_1, \cdots, \bm{h}_{M+N+2}, \bm{h}_{M+N+3}]$. Then we split the representations for two arguments:
\begin{align}
    [\bm{h}_0^1, \bm{h}_1^1, \cdots, \bm{h}_{M+1}^1] &= [\bm{h}_0, \bm{h}_1, \cdots, \bm{h}_{M+1}] \nonumber \\
    [\bm{h}_0^2, \bm{h}_1^2, \cdots, \bm{h}_{N+1}^2] &= [\bm{h}_{M+2}, \bm{h}_{M+3}, \cdots, \bm{h}_{M+N+3}].
\end{align}

\subsection{Bilateral Multi-Perspective Matching}
The resulting contextualized representations are sent to the matching module. 
Those complex discourse relations usually cannot be easily derived from surface features from argument pairs so that a multiple-perspective matching is required.
We follow \cite{Wang2017Bilateral} to utilize full-matching, maxpooling-matching, attentive-matching, and max-attentive-matching but reduce parameters by weight sharing.

We first define a multi-perspective cosine similarity:
\scalebox{0.88}{\parbox{1.14\linewidth}{
\begin{align}
    \bm{c} =& \text{MultiCos}(\bm{v}^1, \bm{v}^2; \bm{W}^c)  \nonumber \\
    =& [
    \cos(\bm{W}_{1,:}^c \circ \bm{v}^1, \bm{W}_{1,:}^c \circ \bm{v}^2), 
    \cdots,
    \cos(\bm{W}_{l,:}^c \circ \bm{v}^1, \bm{W}_{l,:}^c \circ \bm{v}^2)],
\end{align}
}}
where $\bm{c} \in \mathbb{R}^{l}$ is the similarity vector, $\bm{W}^c \in \mathbb{R}^{l \times d}$ is a trainable matrix, $\bm{v}^1, \bm{v}^2 \in \mathbb{R}^{d}$, and $d$ is the dimension of $\bm{v}^1$, $\bm{v}^2$ and $l$ is the number of perspectives.

And then, we compute four matching results based on the above multi-perspective cosine similarity.

\subsubsection{Full-Matching} 
We compute the full matching vector for each token in both arguments as follows:

\scalebox{0.88}{\parbox{1.09\linewidth}{
\begin{align}
    {\bm{\hat m}}_i^1 &= [\text{MultiCos}(\bm{h}_i^1, \bm{h}_0^2; \bm{\hat W}^f), \text{MultiCos}(\bm{h}_i^1, \bm{h}_{N+1}^2; \bm{\hat W}^l)] \nonumber \\
    {\bm{\hat m}}_j^2 &= [\text{MultiCos}(\bm{h}_0^1, \bm{h}_j^2; \bm{\hat W}^f), \text{MultiCos}(\bm{h}_{M+1}^1, \bm{h}_j^2; \bm{\hat W}^l)],
\end{align}
}}
where $\bm{\hat W}^f, \bm{\hat W}^l \in \mathbb{R}^{l \times d}$ are trainable matrices for the first token and the last token respectively, ${\bm{\hat m}}_i^1 \in \mathbb{R}^{2l}$ is the full-matched vector for $i^{th}$ word in $Arg_1$ and ${\bm{\hat m}}_j^2 \in \mathbb{R}^{2l}$ is for $j^{th}$ word in $Arg_2$.
As we add special tokens for each argument (\textit{CLS} and \textit{SEP} for $Arg_1$, \textit{SEP} and \textit{EOS} for $Arg_2$) and these special tokens contain much semantic information of the whole argument, we use these special tokens to consider the full matching in additional to the individual tokens.

\subsubsection{Maxpooling-Matching}
The maxpololing-matching is similar to the full-matching but more focused on the most important part. 
We need to compute all matching results and select the maximum elements.
\begin{align}
    {\bm{\check m}}_i^1 &= \max_{0\leq j \leq N+1} \text{MultiCos}(\bm{h}_i^1, \bm{h}_j^2; \bm{\check W}) \nonumber \\
    {\bm{\check m}}_j^2 &= \max_{0\leq i \leq M+1} \text{MultiCos}(\bm{h}_i^1, \bm{h}_j^2; \bm{\check W}),
\end{align}
where $\bm{\check W} \in \mathbb{R}^{l \times d}$ is a trainable matrix for the maxpooling-matching and $\max$ is the element-wise maximum.

\subsubsection{Attentive-Matching}
The attentive-matching uses the idea of the weighted mean.
We need to compute weights based on the cosine similarity and compute the weighted average vectors:
\begin{align}
    \bm{\bar h}^1 &= \frac{\sum_{0 \leq i \leq M+1}{c_{i,j}\cdot \bm{h}_i^1}}{\sum_{0 \leq i \leq M+1}{c_{i,j}}} \nonumber \\
    \bm{\bar h}^2 &= \frac{\sum_{0 \leq j \leq N+1}{c_{i,j}\cdot \bm{h}_j^2}}{\sum_{0 \leq j \leq N+1}{c_{i,j}}},
\end{align}
where $c_{i,j} = \cos(\bm{h}_i^1, \bm{h}_j^2)$ is the cosine similarity of $\bm{h}_i^1$ and $\bm{h}_j^2$.
Based on the weighted average representations of $Arg_1$ and $Arg_2$, we can compute the corresponding attentive vectors:
\begin{align}
    {\bm{\bar m}}_i^1 &= \text{MultiCos}(\bm{h}_i^1, \bm{\bar h}^2; \bm{\bar W}) \nonumber \\
    {\bm{\bar m}}_j^2 &= \text{MultiCos}(\bm{h}_i^2, \bm{\bar h}^1; \bm{\bar W}),
\end{align}
where $\bm{\bar W} \in \mathbb{R}^{l \times d}$ is a trainable matrix for the attentive-matching.

\subsubsection{Max-Attentive-Matching}
Instead of taking the weighted average of all the contextual embeddings as the attentive vector, the max-attentive-matching picks the contextual embedding with the highest cosine similarity as the attentive vector:
\begin{align}
    {\bm{\tilde m}}_i^1 &= \text{MultiCos}(\bm{h}_i^1, \bm{\tilde h}^2; \bm{\tilde W}) \nonumber \\
    {\bm{\tilde m}}_j^2 &= \text{MultiCos}(\bm{h}_i^2, \bm{\tilde h}^1; \bm{\tilde W}) ,
\end{align}
where $\bm{\tilde W} \in \mathbb{R}^{l \times d}$ is a trainable matrix for the max-attentive-matching, $\bm{\tilde h}^1$ corresponds to the contextualized representation $\bm{h}_i^1$ with $\mathop{\arg\max}_{0 \leq i \leq M+1}{c_{i,j}}$, and $\bm{\tilde h}^2$ corresponds to that of $\bm{h}_j^2$ with $\mathop{\arg\max}_{0 \leq j \leq N+1}{c_{i,j}}$.

Finally, the bilateral multi-perspective matching results can be obtained by concatenating the four matching results:
\begin{align}
    {\bm{m}}_i^1 &= [{\bm{\hat m}}_i^1, {\bm{\check m}}_i^1, {\bm{\bar m}}_i^1, {\bm{\tilde m}}_i^1] \nonumber \\
    {\bm{m}}_j^2 &= [{\bm{\hat m}}_j^2, {\bm{\check m}}_j^2, {\bm{\bar m}}_j^2, {\bm{\tilde m}}_j^2],
\end{align}
where ${\bm{m}}_i^1, {\bm{m}}_j^2 \in \mathbb{R}^{5l}$.

\subsection{Gated Information Fusion}
The matching results contain the interaction between two arguments but miss the interaction inside each argument.
The most direct method is to apply self-attention on the concatenation of $\bm{h}$ and $\bm{m}$.
Multi-head attention with layer normalization has been used in the Transformer architecture~\cite{Vaswani2017Attention}.
We modify it as a gated multi-head attention without layer normalization to remain the scale of each dimension and retain the non-linear capability:

\scalebox{0.91}{\parbox{1.05\linewidth}{
\begin{align}
    \bm{Q}' = \text{Multi}&\text{Head}(\bm{Q}, \bm{K}, \bm{V}) \\
    a = \text{sigmo}&\text{id}({\bm{W}^a}^\top[\bm{Q}, \bm{Q}']) \\
    \text{GatedMultiHead}(\bm{Q}, \bm{K}, \bm{V}) &= a \odot \bm{Q}' + (1-a) \odot \bm{Q},
\end{align}
}}
where $\bm{Q} \in \mathbb{R}^{d_q}, \bm{K} \in \mathbb{R}^{d_k}, \bm{V} \in \mathbb{R}^{d_v}, \bm{Q}'\in \mathbb{R}^{d_q}$ are the query, key, value, and output respectively, 
$\bm{W}^a \in \mathbb{R}^{2d_q \times 1}$ is trainable matrix to transform the concatenation of $\bm{Q}$ and $\bm{Q}'$ to a scalar value $a$. The final attention result is controlled by the gate $a$.

To fuse the contextualized representations $[\bm{h}_i^1, \cdots, \bm{h}_{M+1}^1]$ and matching vectors $[\bm{m}_i^1, \cdots, \bm{m}_{M+1}^1]$, we apply the gated multi-head attention over the concatenation results:
\scalebox{0.952}{\parbox{1.05\linewidth}{
\begin{align}
\label{eq:gated_fusion1}
    \bm{Q}_{i,:}^1 = &\ \bm{K}_{i,:}^1 = \bm{V}_{i,:}^1 = [\bm{h}_i^1, \bm{m}_i^1] \\
\label{eq:gated_fusion2}
    [\bm{f}_1^1, \bm{f}_2^1, \cdots, \bm{f}_{M+1}^1] &= \text{GatedMultiHead}(\bm{Q}^1, \bm{K}^1, \bm{V}^1).
\end{align}
}}
We can also get $[\bm{f}_1^2, \cdots, \bm{f}_{N+1}^2]$ similarly.

\subsection{Aggregation and Prediction}

The aggregation layer aggregates the two sequences of fused vectors into a fixed-length matching vector.
Inspired by \cite{Kim2016Character}, we use a convolutional neural network followed by a highway network to capture unigram, bigram, ..., $n$-gram information.
Assuming we apply $z$ convolutional operations $\text{Conv}_1$, $\text{Conv}_2$, ..., $\text{Conv}_z$ followed by max pooling operations to $[\bm{f}_{i}^1, \cdots, \bm{f}_{M+1}^1]$, then the output of each conv-pool operation applied on arguments is
\scalebox{0.952}{\parbox{1.05\linewidth}{
\begin{align}
    \bm{u}_c^1 = \max_{0 \leq i \leq M+1-c}{\text{ReLU}(\text{Conv}_c([\bm{f}_{i}^1, \cdots, \bm{f}_{i+c}^1]))},
\end{align}
}}
where the kernel size of $\text{Conv}_c$ is $c$, the stride is 1, and the dimension of $\bm{u}_c^1$ depends on the number of filters (assume we use $s$ filters).

The flatten concatenation of all conv-pool results are sent as the input of a highway network to extract more features:
\begin{align}
    \bm{u}^1 &= \text{Flatten}([\bm{u}_1^1, \bm{u}_2^1, \cdots, \bm{u}_z^1]) \\
    \bm{u}^{1'} &= \text{ReLU}({\bm{W}^h}^\top\bm{u}^1) \\
    g^1 &= \text{sigmod}({\bm{W}^g}^\top\bm{u}^1) \\
    \bm{o}^1 &= g^1 \odot \bm{u}^{1'} + (1-g^1) \odot \bm{u}^1,
\end{align}
where $\bm{u}^1 \in \mathbb{R}^{z s}$ is a flatten vector, $\bm{W}^h \in \mathbb{R}^{z s \times z s}, \bm{W}^g \in \mathbb{R}^{z s \times 1}$ are trainable.
As before, we can get $\bm{o}^2$ similarly.

Finally, the prediction layer evaluates the probability distribution $\text{Pr}(y|Arg_1, Arg_2)$. 
We use a two layer feed-forward neural network to consume $\bm{o}^1$ and $\bm{o}^2$, and apply the \text{softmax} function in the last layer.
To approximate the data distribution, we optimize cross-entropy loss between the outputs of the prediction layer and the ground-truth labels.\footnote{Some instances in PDTB and CoNLL were annotated with more than one label. Following previous work and the scorer at \url{https://github.com/attapol/conll16st}, a prediction is regarded as correct once it matches one of the ground-truth labels.}

\begin{table}[t]
\small
\centering
\begin{tabular}{l|c|c|c}
\toprule
Relation &  training & validation & test \\
\midrule
Comp. & 1,942 & 197 & 152 \\
Cont. & 3,342 & 295 & 279 \\
Exp. & 7,004 & 671 & 574 \\
Temp. & 760 & 64 & 85 \\
\hline
Total & 12,362 & 1,183 & 1,046\\
\bottomrule
\end{tabular}
\caption{Statistics of four top level implicit senses in PDTB 2.0.}
\label{table:pdtb_statistics}
\end{table}

\begin{table*}[t]
\small
\centering
\begin{tabular}{l|cc|c|c|c}
\toprule
\multicolumn{1}{c|}{Model}
& \multicolumn{2}{c|}{PDTB-4 (F1;Acc)} & PDTB-11 (Acc) & CoNLL-Test (Acc) & CoNLL-Blind (Acc) \\
\midrule
\cite{Ji2015One} & - & - & 44.59 & - & - \\
\cite{Liu2016Implicit} & 44.98 & 57.27 & - & - & - \\
\cite{Liu2016Recognizing}\footnotemark[2] & 46.29 & 57.57 & - & - & - \\
\cite{Rutherford2016Robust} & - & - & - & 40.91 & 34.20 \\
\cite{Lan2017Multi} & 47.80 & 57.39 & - & 39.40 & 40.12 \\
\cite{Bai2018Deep} & 51.06 & - & 48.22 & - & - \\
\cite{Xu2018Using} & 44.48 & 60.63 & - & - & -  \\
\cite{Dai2018Improving} & 48.82 & 57.44 & - & - & - \\
\cite{Dai2019A} & 52.89 & 59.66 & 48.23 & - & - \\
\cite{Shi2019Next}\footnotemark[2] & - & - & 53.23 (0.39) & - & - \\
\cite{Varia2019Discourse}\footnotemark[3] & 50.20 & 59.13 & - & 39.39 & 39.36 \\
\hline
BMGF-RoBERTa-Siamese & 53.70 (0.31) & 62.27 (0.67) & 50.37 (0.59) & 47.42 (1.04) &  46.24 (1.07) \\
BMGF-RoBERTa & \bf 63.39 (0.56) & \bf 69.06 (0.43) & \bf 58.13 (0.67) & \bf 57.26 (0.75) & \bf 55.19 (0.55) \\
\bottomrule
\end{tabular}
\caption{Performance of multi-class classification on PDTB and CoNLL-2016 in terms of accuracy (Acc) (\%) and macro-averaged F1 (F1) (\%) and corresponding standard deviation.}
\label{table:multiclass}
\end{table*}

\begin{table}[t]
\small
\centering
\begin{tabular}{l|c|c|c|c}
\toprule
\multicolumn{1}{c|}{Model}
& Comp. & Cont. & Exp. & Temp. \\
\midrule
\cite{Ji2015One} & 35.93 & 52.78 & - & 27.63 \\
\cite{Rutherford2015Improving} & 41.00 & 53.80 & 69.40 & 33.30 \\
\cite{Liu2016Implicit}  & 37.91 & 55.88 & 69.97 & 37.17 \\
\cite{Liu2016Recognizing}\footnotemark[2] & 39.86 & 54.48 & 70.43 & 38.84 \\
\cite{Qin2016A} & 38.67 & 54.91 & 71.50 & 32.76 \\
\cite{Lan2017Multi}  & 40.73 & 58.96 & 72.47 & 38.50 \\
\cite{Bai2018Deep} & 47.85 & 54.47 & 70.60 & 36.87 \\
\cite{Dai2018Improving} & 46.79 & 57.09 & 70.41 & 45.61 \\
\cite{Varia2019Discourse}  & 44.10 & 56.02 & 72.11 & 44.41 \\
\hline
BMGF-RoBERTa-Siamese &  44.19 & 56.36 & 72.90 & 39.65  \\
BMGF-RoBERTa & \bf 59.44 & \bf 60.98 & \bf 77.66 & \bf 50.26 \\
\bottomrule
\end{tabular}
\caption{Performance of multiple binary classification on the top level classes in PDTB in terms of F1 (\%).}
\label{table:binary}
\end{table}
\section{Experiments}
In this section, we show our experimental results and detailed analysis of different representation learning modules.
\subsection{Datasets}
\paragraph{The Penn Discourse Treebank 2.0 (PDTB 2.0)}
PDTB 2.0 is a large scale corpus annotated with information related to discourse structure and discourse semantics, containing 2,312 Wall Street Journal (WSJ) articles.
PDTB 2.0 has three levels of senses, i.e., classes, types, and sub-types.
Most work focuses on the top level, including \textit{Comparison} (\textit{Comp.}), \textit{Contingency} (\textit{Cont.}), \textit{Expansion} (\textit{Exp.}), and \textit{Temporal} (\textit{Temp.}).
We follow \cite{Ji2015One} to split section 2-20, 0-1, and 21-22 as training, validation, and test sets respectively.
We not only evaluate our model on the four top-level implicit senses but the 11 major second level implicit types.
Table~\ref{table:pdtb_statistics} shows the statistics of the top-level senses.

\paragraph{The CoNLL-2016 Shared Task (CoNLL)}
The CoNLL-2016 shared task~\cite{Xue2016CoNLL} provides more abundant annotation for shadow discourse parsing.
The PDTB section 23 (CoNLL-Test) and newswire texts (CoNLL-Blind) following the PDTB annotation guidelines were organized as the test sets.
CoNLL merges several labels. For example, \textit{Contingency.Pragmatic cause} is
merged into \textit{Contingency.Cause.Reason} to remove the former type with very few samples.
Finally, there is a flat list of 15 sense categories to be classified.

\subsection{Parameter Settings}
We use the RoBERTa-base as our context representation layer. Because the data size is not large enough so that we freeze all parameters but add trainable segment embeddings like BERT.
The number of perspectives of the matching layer $l$, the number of heads of the fusion layer $h$, the number of convolutional operations $z$, the number of filters $s$ are set as 16, 16, 2, 64 respectively, and all hidden dimensions are set as 128.
We apply dropout to every layer with the dropout rate 0.2, clip the gradient L2-norm with a threshold 2.0, and add the L2 regularization with coefficient 0.0005 to alleviate overfitting.
We adopt Adam~\cite{Kingma2014Adam} with an initial learning rate of 0.001 and a batch size of 32 to train models up to 50 epochs.
All experiments are performed three times with $2 \times 32$ GB NVIDIA V100 GPUs and all reported results are averaged performance. Our model and variants are all implemented by PyTorch.

\subsection{Experimental Results}
\footnotetext[2]{We select the best results from their different settings.}
\footnotetext[3]{We use the average results of their best single model.}

We compare {\bf BMGF-RoBERTa} with other recent state-of-the-art baselines on multi-class classification and multiple binary classification to demonstrate the importance of our representation, matching, and fusion modules.
A siamese network (denoted as {\bf BMGF-RoBERTa-Siamese}) whose encoder does not concatenate embeddings in two arguments but uses embeddings from RoBERTa separately obtained from each argument is also implemented for analysis.

Macro-averaged F1 is the main metric for the four-class classification, and accuracy is considered as the main metric for remaining multi-class classification problems. 
For binary classification, F1 is adopted to evaluate the performance on each class.

\begin{table*}[t]
\small
\centering
\begin{tabular}{l|cc|c|c|c}
\toprule
\multicolumn{1}{c|}{Model}
& \multicolumn{2}{c|}{PDTB-4 (F1;Acc)} & PDTB-11 (Acc) & CoNLL-Test (Acc) & CoNLL-Blind (Acc) \\
\midrule
BMGF-RoBERTa & \bf 63.39 & \bf 69.06 & \bf 58.13 & \bf 57.26 & \bf 55.19 \\
\quad w/o SE & 55.15 & 63.29 & 50.72 & 49.67 & 48.06 \\
\quad w/o GF & 60.40 & 66.28 & 55.31 & 53.92 & 50.51 \\
\quad w/o BM & 61.93 & 68.83 & 57.23 & 56.22 & 54.39  \\
\quad w/o SE,GF & 53.36 & 62.24 & 50.14 & 48.07 & 46.56 \\
\quad w/o SE,BM & 51.42 & 63.22 & 50.37 & 47.51 & 48.22 \\
\quad w/o BM,GF & 59.32 & 66.89 & 55.89 & 55.22 & 52.18 \\
\quad w/o SE,BM,GF (RoBERTa) & 50.97 & 62.11 & 49.76 & 47.33 & 46.16 \\
\hline 
RoBERTa w/ SourceAttn & 54.42 & 62.24 & 49.79 & 47.85 & 46.16 \\
RoBERTa w/ SourceAttn, SelfAttn & 52.89 & 62.56 & 50.02 & 47.94 & 47.59 \\
\bottomrule
\end{tabular}
\caption{Architecture ablation analysis on multi-class classification.}
\label{table:arch_ablation}
\end{table*}

\subsubsection{Multi-class Classification}
We first evaluate our model on multi-class classification which is the final part of a shallow discourse parser as usual.
Two different settings are set for PDTB in view of the hierarchical categories.
One is the four-class classification (denoted as PDTB-4) at the top-level and the other is the 11-class classification (denoted as PDTB-11) at the second level.
We conduct 15-class classification for CoNLL dataset, denoted as CoNLL-Test and CoNLL-Blind for the test set and blind set.

Results of our model and baselines are shown in Table~\ref{table:multiclass}, and BMGF-RoBERTa achieves the new state-of-the-art performance with substantial improvements on all implicit discourse classification settings.
Most previous systems considered siamese architectures except \cite{Shi2019Next} which was the first to consider two arguments as a whole and fine-tune BERT based on the next sentence prediction loss for discourse relation classification.
Our variant model BMGF-RoBERTa-Siamese can also beat other siamese models and narrow 40\% of the gap with \cite{Shi2019Next}.
Furthermore, BMGF-RoBERTa is the first model that exceeds the 60\% macro-averaged F1  and approximates 70\% accuracy in the four-class classification on PDTB.

\begin{table}[t]
\small
\centering
\begin{tabular}{l|c|c|c|c}
\toprule
\multicolumn{1}{c|}{Model}
& Comp. & Cont. & Exp. & Temp. \\
\midrule
BMGF-RoBERTa & \bf 59.44 & \bf 60.98 & \bf 77.66 & \bf 50.26 \\
BMGF-RoBERTa w/o SE & 47.08 & 55.64 & 73.26 & 39.91 \\
BMGF-RoBERTa-Siamese &  44.19 & 56.36 & 72.90 & 39.65 \\
\bottomrule
\end{tabular}
\caption{Encoder ablation analysis on multiple binary classification.}
\label{table:encoder_ablation}
\end{table}

\subsubsection{Binary Classification}
To compare previous results more directly and analyze model behaviour more comprehensively, we conduct experiments based on one-vs-rest binary classifiers on the four top-level classes in PDTB.
Table~\ref{table:binary} shows the results of binary classification on the top-level classes in PDTB. 
It is as expected that BMGF-RoBERTa has the best performance, but it is still surprising that the improvement on \textit{Comp.} exceeds the previous best more than 11\% and the improvement on \textit{Temp.} exceeds the previous best around 5\%.
In contrast, the performance of the siamese variant looks a bit ordinary.
Results of binary classification indicate that RoBERTa encoder can further capture the deeper discourse relations after giving the ``previous-next'' context relationship.
Such hard discourse relations can not be handled well when two arguments are encoded separately.
However, it is still unclear whether the self-attention during encoding or segment embeddings provides the most useful information.
We discuss more in Section~\ref{sec:ablation}.

\subsection{Ablation Study}
\label{sec:ablation}

We first analyze the effectiveness of the encoding module, the matching module, and the fusion module.
To this end, we remove the segment embeddings (SE), the bilateral multi-perspective matching (BM), and the gated information fusion (GF) from BMGF-RoBERTa one by one.
To compare the common matching and fusion modules, we implement another two ablation models based on the original RoBERTa:
{\bf RoBERTa w/ SourceAttn} is a model that uses the RoBERTa to encode the concatenation of two arguments and then do the source attention between two separated contextualized embeddings with the residual connection; {\bf RoBERTa w/ SourceAttn, SelfAttn} adds the self-attention and the residual connection after the source attention.

Table~\ref{table:arch_ablation} shows the results. We can see that eliminating any of the three modules would hurt the performance significantly.
The original RoBERTa model cannot surpass the previous best siamese network on PDTB-4 too much or outperform BERT on PDTB-11, but the SE, BM, and GF can help it improve around 9.3\% on average.
At the same time, the segment embeddings improve mostly.
When comparing source attention and self-attention with our bilateral multi-perspective matching and gated information fusion, we can also conclude the combination of BM and GF is much better than those vanilla multi-head attention mechanisms when BMGF-RoBERTa w/o SE as the reference.
Moreover, the self-attention hurts the F1 score in PDTB-4.

BMGF-RoBERTa-Siamese shown in Table~\ref{table:multiclass} is not comparable with BMGF-RoBERTa w/o SE and show in Table~\ref{table:arch_ablation}. 
Separately encoding cannot benefit from the longer self-attention after concatenation.
And we can also find SE can improve much more in Table~\ref{table:arch_ablation}, indicating that the longer self-attention is not as straightforward as segment embeddings.

We further design the encoder ablation study for the encoder module and results are listed in Table~\ref{table:encoder_ablation}.
We can find that the main gaps between encoding without concatenating (BMGF-RoBERTa-Siamese) and encoding without segment embeddings (BMGF-RoBERTa w/o SE) come from the \textit{Comp.} and \textit{Temp.} classes.
And adding SE brings a qualitative leap in all four classes because BMGF-RoBERTa finally captures clues of two arguments by multiple attention and relative segment order embeddings.
That is, a more powerful representation layer can bring unexpected improvement and understand implicit relations better.
But how to inject useful information to the representation module looks more important.
In our paper, we inject the ``previous-next'' context relationship by the concatenation and segment embeddings for implicit discourse relation classification.
\section{Conclusion}

We present a novel model, BMGF-RoBERTa, that combines representation, matching, and fusion modules for implicit discourse relation classification.
Experimental results show BMGF-RoBERTa outperforms all previous state-of-the-art systems with substantial margins on the PDTB dataset and CoNLL datasets.
We find the ``previous-next'' context relationship provides much information for this difficult task.
Furthermore, the effectiveness of these well-designed modules and what is important in the representation layer are thoroughly analyzed in the ablation study.
One important future work would be to find more clues for better representations within and even beyond the word- and sentence-levels.
\section*{Acknowledgements}
This paper was supported by the Early Career Scheme (ECS, No. 26206717) from the Research Grants Council in Hong Kong and the Gift Fund from Huawei Noah's Ark Lab.


{\small
\bibliographystyle{named}
\bibliography{ijcai20}
}
\end{document}